\pdfoutput=1
\pdfoutput=1
\documentclass[sigconf, authorversion]{acmart}

\usepackage[table]{xcolor}
\definecolor{hotpink}{rgb}{1,0.41,0.71}

\usepackage{hyperref}
\usepackage{url}
\usepackage[most]{tcolorbox}
\usepackage{float}
\usepackage{threeparttable}
\usepackage{graphicx}

\usepackage{xcolor}
\usepackage{textcomp}

\AtBeginDocument{%
  }

\setcopyright{acmlicensed}
\copyrightyear{2025}
\acmYear{2025}
\acmDOI{XXXXXXX.XXXXXXX} 
\acmConference[FinAI '25]{Advances in Financial AI}{November 14, 2025}{Seoul, Republic of Korea}
\acmBooktitle{Proceedings of the 34th ACM International Conference on Information and Knowledge Management (CIKM '25), November 10--14, 2025, Seoul, Republic of Korea}
\acmISBN{978-1-4503-XXXX-X/2025/11} 

\begin{document}

\title{GuruAgents: Emulating Wise Investors with Prompt-Guided LLM Agents}

\author{Yejin Kim}
\authornotemark[1]
\email{yejin.kim.ds@meritz.com}
\affiliation{%
  \institution{Meritz Fire \& Marine Insurance}
  \country{Republic of Korea}
  }
\affiliation{%
  \institution{AI Quant Lab, MODULABS}
  \city{Seoul}
  \country{Republic of Korea}}

\author{Youngbin Lee}
\authornote{These authors contributed equally to this work.}
\email{youngandbin@elicer.com}
\affiliation{%
  \institution{Elice}
  \country{Republic of Korea}
  }
  
\affiliation{%
  \institution{AI Quant Lab, MODULABS}
  \city{Seoul}
  \country{Republic of Korea}}

\author{Juhyeong Kim}
\email{juhyeong.kim@miraeasset.com}
\email{nonconvexopt@gmail.com}
\affiliation{%
  \institution{Mirae Asset Global Investments}
  \country{Republic of Korea}
  }
\affiliation{%
  \institution{AI Quant Lab, MODULABS}
  \city{Seoul}
  \country{Republic of Korea}}

\author{Yongjae Lee}
\authornote{Corresponding author}
\email{yongjaelee@unist.ac.kr}
\affiliation{%
  \institution{Ulsan National Institute of Science and Technology}
  \city{Ulsan}
  \country{Republic of Korea}}

\renewcommand{\shortauthors}{Kim et al.}

\begin{abstract}
This study demonstrates that GuruAgents, prompt-guided AI agents, can systematically operationalize the strategies of legendary investment gurus. We develop five distinct GuruAgents, each designed to emulate an iconic investor, by encoding their distinct philosophies into LLM prompts that integrate financial tools and a deterministic reasoning pipeline. In a backtest on NASDAQ-100 constituents from Q4 2023 to Q2 2025, the GuruAgents exhibit unique behaviors driven by their prompted personas. The Buffett GuruAgent achieves the highest performance, delivering a 42.2\% CAGR that significantly outperforms benchmarks, while other agents show varied results. These findings confirm that prompt engineering can successfully translate the qualitative philosophies of investment gurus into reproducible, quantitative strategies, highlighting a novel direction for automated systematic investing. The source code and data are available at \textcolor{hotpink}{\url{https://github.com/yejining99/GuruAgents}}.
\end{abstract}





\maketitle

\section{Introduction}

The strategies of legendary investors are often codified into clear, memorable rules. These frameworks emphasize principles such as margin of safety \cite{graham2003intelligent}, accounting-based diagnostics \cite{piotroski2000value, altman2010corporate}, and disciplined portfolio construction \cite{greenblatt2005little}. Despite their clarity, however, operationalizing such strategies in a systematic and generalizable manner remains challenging. Translating the qualitative philosophies of these gurus into deterministic, data-driven rules often requires expert judgment, leaving a gap between the conceptual elegance of their doctrines and consistent automation.

Recent progress in LLMs offers a promising bridge for this gap. LLMs have demonstrated the ability to follow structured role instructions and adopt coherent personas, allowing them to emulate decision-making styles when guided by carefully engineered prompts \cite{Li2023CAMEL}. Beyond persona construction, LLMs acting as autonomous agents have been shown to plan, reason, and interact with environments through tool use \cite{liu2023agentbench,Yao2023ReAct,Schick2023Toolformer,Qin2023ToolLLM,Xu2023ReWOO}. These capabilities highlight the potential of LLM-based agents to not only interpret financial data but also to operationalize established investment philosophies with transparency and reproducibility.

The application of LLMs to the financial domain is a rapidly emerging field of research. Recent studies primarily focus on using LLMs for quantitative tasks, such as analyzing the sentiment of financial news to forecast stock returns \cite{lopez2023can} or extracting information from dense analyst reports \cite{kim2023llms}. While specialized models like BloombergGPT enhance these data-processing capabilities \cite{wu2023bloomberggpt}, and other works explore automated signal generation \cite{yang2024finrobot} and integration into optimization frameworks \cite{lee2025integrating}, these approaches treat LLMs as sophisticated tools for pattern recognition. They leave unexplored the possibility of capturing the qualitative, principle-driven wisdom that underpins the long-term success of legendary investors.

To address this limitation, we introduce GuruAgents: prompt-guided LLM agents designed to emulate the distinct philosophies and decision-making personas of wise investors. This study investigates whether these GuruAgents, guided solely by prompt engineering and tool integration, can faithfully translate the qualitative doctrines of investment masters into quantitative and reproducible portfolio decisions. By encoding role definitions, canonical quotations, and deterministic scoring rules directly into system prompts, we explore the extent to which LLMs can serve as faithful proxies for the strategic minds of investment gurus.

\section{Methodology}


This study introduces GuruAgents, a system of prompt-guided LLM agents designed to emulate and operationalize the strategies of renowned investment gurus. Each GuruAgent internalizes the unique investment philosophy and decision-making framework of a specific guru through carefully crafted prompt engineering. This section describes the prompt engineering framework that underpins the construction of these GuruAgents, as well as the detailed implementation of each one.

\subsection{Prompt Engineering Framework}

To build effective LLM-based investment agents, we developed a prompt engineering framework consisting of three core components: \textit{role-based persona construction}, \textit{tool integration design}, and a \textit{deterministic reasoning pipeline}. This framework enables the agents to interpret complex financial data, execute investment strategies, and ensure transparency in decision-making.

\subsubsection{Role-Based Persona Construction}

Each investment agent is assigned a clear persona corresponding to a specific investor, ensuring faithful reproduction of that investor’s philosophy and methods. This approach plays a crucial role in maintaining consistency in agent behavior and decision-making.

\begin{itemize}
    \item \textbf{Role definition}: Prompts begin with instructions such as \textit{``You are \{Investor\}, \{core philosophy\} ...''}, explicitly assigning the agent a role.
    \item \textbf{Codification of beliefs}: Each prompt embeds the investor’s well-known principles and maxims, ensuring that decisions reflect the underlying philosophy.
    \item \textbf{Tone and manner}: The prompts also incorporate the distinctive voice and demeanor of the investor—for example, Graham’s prudent and skeptical tone versus Buffett’s patient and plainspoken style.
\end{itemize}

\subsubsection{Tool Integration Design}

Accurate financial analysis is essential for agents to make meaningful investment decisions. To this end, we integrated a suite of computational tools directly into the prompts, allowing agents to actively leverage structured financial metrics before portfolio construction.

\begin{itemize}
    \item \textbf{Standard financial metrics}: They provide standardized measures of liquidity, profitability, and leverage.
    \item \textbf{Valuation tools}: They compute price multiples, market capitalization, net current asset value (NCAV), and net-net status, supporting equity valuation.
    \item \textbf{Strategy-specific tools}: They include additional functions that reflect the idiosyncratic approaches of particular investors, such as Piotroski’s F-Score signals or Greenblatt’s Magic Formula components. The agents rely on these tool outputs for the quantitative evaluation of firms.
\end{itemize}

\subsubsection{Deterministic Reasoning Pipeline}

To minimize uncontrolled variability and ensure reproducibility, we established a \textit{deterministic reasoning pipeline}. Each agent adheres to the following fixed sequence, guaranteeing that identical inputs always yield identical outputs.

\begin{enumerate}
    \item \textbf{Metric collection}: Predefined functions are invoked to extract the required financial indicators.
    \item \textbf{Scoring}: Firms are scored according to investor-specific weighting schemes and penalty/bonus adjustments. The scoring process is explicitly algorithmic, ensuring deterministic results.
    \item \textbf{Portfolio construction}: Scores are converted into portfolio weights, normalized to sum to 100. In the case of ties, a predefined priority order—liquidity ratio, debt ratio, profit margin—is applied to break ties consistently.
\end{enumerate}

The final output is standardized into a table with four columns: \textit{Ticker, Score, Weight (\%), and Reason}. Scores are reported to two decimal places, weights as percentage integers summing to 100, and each stock is accompanied by a one-sentence rationale grounded in its metrics. This standardization allows the outputs to be directly utilized in backtesting and comparative evaluation.

\begin{figure*}[!ht]
    \centering
    \includegraphics[width=\textwidth]{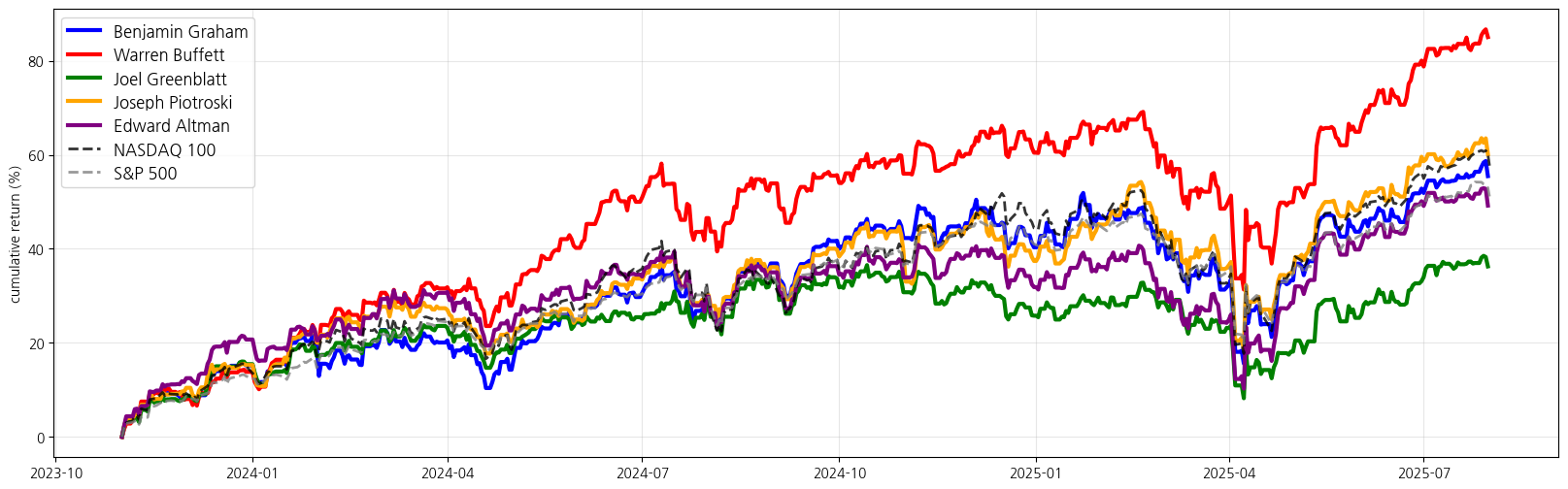}
    \caption{Cumulative returns of five legendary investor-inspired agents (Graham, Buffett, Greenblatt, Piotroski, Altman) and benchmarks (NASDAQ 100, S\&P 500) from Q4 2023 to Q2 2025.}
    \Description{Line chart showing cumulative returns of five investor-inspired agents and benchmarks (NASDAQ 100, S\&P 500) from Q4 2023 to Q2 2025.}
    \label{fig:cumulative}
\end{figure*}

\begin{table*}[!ht]
\centering
\caption{Summary performance metrics of agents and benchmarks.}
\label{tab:performance}
\setlength{\tabcolsep}{6pt}
\renewcommand{\arraystretch}{1.2}
\resizebox{\textwidth}{!}{%
\begin{tabular}{lrrrrrrrrrr}
\hline
\textbf{Strategy} & \textbf{CAGR\(\uparrow\)} & \textbf{mean (daily)\(\uparrow\)} & \textbf{std (daily)\(\downarrow\)} & \textbf{mean (ann.)\(\uparrow\)} & \textbf{std (ann.)\(\downarrow\)} & \textbf{Sharpe\(\uparrow\)} & \textbf{Sharpe (ann.)\(\uparrow\)} & \textbf{MDD\(\uparrow\)} & \textbf{VaR$_{0.9}$\(\uparrow\)} & \textbf{CVaR$_{0.9}$\(\uparrow\)} \\
\hline
\multicolumn{11}{l}{\emph{Legendary Investors}} \\
Benjamin Graham   & 28.7401 & 0.0008 & 0.0119 & 0.1921 & 0.1896 & 0.0638 & 1.0132 & -23.8873 & -1.0563 & -2.1079 \\
Warren Buffett    & \textbf{42.2341} & \underline{0.0010} & 0.0117 & \underline{0.2603} & 0.1860 & \underline{0.0881} & \underline{1.3991} & -22.3440 & \textbf{-0.8934} & -1.9950 \\
Joel Greenblatt   & 19.3799 & 0.0005 & \textbf{0.0098} & 0.1342 & \textbf{0.1551} & 0.0545 & 0.8652 & \underline{-20.7409} & -0.9877 & \textbf{-1.7126} \\
Joseph Piotroski  & \underline{30.9300} & 0.0008 & 0.0111 & 0.2014 & 0.1762 & 0.0720 & 1.1432 & -23.0692 & -1.0250 & -1.9732 \\
Edward Altman       & 25.7406 & 0.0007 & 0.0114 & 0.1744 & 0.1817 & 0.0605 & 0.9598 & -21.7132 & -1.1024 & -2.0331 \\
\hline
\multicolumn{11}{l}{\emph{Benchmarks}} \\
NASDAQ 100        & 29.3611 & \textbf{0.0011} & 0.0135 & \textbf{0.2827} & 0.2150 & 0.0828 & 1.3151 & -22.7683 & -1.3911 & -2.4290 \\
S\&P 500          & 26.3131 & \underline{0.0010} & \underline{0.0107} & 0.2500 & \underline{0.1698} & \textbf{0.0928} & \textbf{1.4728} & \textbf{-18.7552} & \underline{-0.9144} & \underline{-1.8389} \\
\hline
\end{tabular}%
}
\end{table*}

\subsection{Investment Agent Design}
\subsubsection{Benjamin Graham Agent}

Benjamin Graham, known as the father of value investing, emphasized intrinsic value, a ``margin of safety,'' and portfolio-level judgment \cite{graham2003intelligent, graham1974future}. 

\begin{itemize}
    \item Key Quotations in Prompt: 
``The individual investor should act consistently as an investor and not as a speculator'', 
``Have a margin of safety'', 
markets that ``advance too far and decline too far'', 
and the principle that ``relatively little stress'' should be placed on forecasting markets, focusing instead on intrinsic value and financial strength.
\end{itemize}

\subsubsection{Edward Altman Agent}

Edward Altman, a finance professor at NYU, is best known for developing the \emph{Z-Score} models, which use accounting ratios to estimate the likelihood of corporate default \cite{altman2000predicting, altman2010corporate}. His framework classifies firms into Safe, Grey, and Distress zones rather than forecasting directly.

\begin{itemize}
    \item Key Concepts in Prompt: Altman introduced the original Z-Score model to classify firms into 
Safe, Grey, and Distress zones using a linear combination of five ratios. He later refined this framework into the Z\textquotesingle-Score, more suitable for non-manufacturers, 
and the Z\textquotedbl-Score, tailored for emerging markets.
\end{itemize}

\subsubsection{Joel Greenblatt Agent}

Joel Greenblatt, hedge fund manager and author of \textit{The Little Book That Beats the Market}, proposed the \textit{Magic Formula} that ranks stocks by two variables—earnings yield and return on capital—and applies a simple rules-based portfolio process \cite{greenblatt2005little, greenblatt2010little}.
\begin{itemize}
\item Key Concepts in Prompt: Greenblatt’s \emph{Magic Formula} ranks firms by two simple metrics—Earnings Yield 
($\approx$ EBIT/Enterprise Value) and Return on Capital ($\approx$ EBIT/(Net Working Capital + Net PPE)). The rules are deliberately simple and judged at the portfolio 
level rather than on single names, with emphasis on current operating performance 
over forecasting; firms with negative EBIT or non-sensical denominators (e.g., EV $\leq 0$) are excluded.
\end{itemize}

\subsubsection{Joseph Piotroski Agent}
Joseph Piotroski, an accounting scholar, introduced the \textit{F-Score} to identify financially strong value stocks by using a nine-signal checklist based on profitability, leverage/liquidity, and operating efficiency \cite{piotroski2000value}.
\begin{itemize}
    \item Key Concepts in Prompt: Piotroski’s \emph{F-Score} applies nine binary signals spanning profitability, leverage/liquidity, 
and operating efficiency \cite{piotroski2000value}. The framework emphasizes recent fundamental 
improvements and accounting quality rather than forecasting, with typical signals including 
ROA $>$ 0, CFO $>$ 0, improving current ratio, no equity issuance, and rising margins or turnover.
\end{itemize}

\subsubsection{Warren Buffett Agent}

Warren Buffett emphasizes purchasing high-quality businesses at fair prices, concentrating on durable moats, conservative balance sheets, and long-term owner-like thinking grounded in intrinsic value \cite{buffett1994letter}.
\begin{itemize}
    \item Key Quotations in Prompt: Buffett emphasized buying ``a wonderful company at a fair price'' rather than the reverse, with the ideal holding period being ``forever''. 
He distinguished between ``price'' and ``value'', urged investors to stay 
within their ``circle of competence'', and advised being ``fearful when 
others are greedy and greedy when others are fearful''. His letters also 
highlight intrinsic value as the discounted cash that can be taken out of the business, and he voiced skepticism toward EBITDA ``chest-thumping''.
\end{itemize}

\section{Experimental Design}

\subsection{Implementation Details}

The GuruAgents presented in this study are implemented using OpenAI's GPT-4o as the core Large Language Model, accessed via its API. The agentic architecture, including tool integration and the stateful, deterministic reasoning pipeline described in the methodology, is structured and executed using the LangChain and LangGraph frameworks. This allows for a robust and reproducible implementation of each GuruAgent's multi-step decision-making process.

\subsection{Dataset}
The empirical analysis uses data from the NASDAQ-100 constituents covering the period 
\textbf{Q4 2023 to Q2 2025}. 
This horizon is chosen to ensure that the testing window lies beyond the knowledge cutoff of the LLM (GPT-4o), 
so that the model cannot trivially memorize historical outcomes.  
The dataset integrates:
\begin{itemize}
    \item \textbf{Market data:} OHLCV prices, number of shares outstanding, and market capitalization.
    \item \textbf{Accounting data:} Quarterly balance sheet (BS), cash flow statement (CF), and income statement (IS).
\end{itemize}

\subsection{Backtesting Framework}

Each GuruAgent's portfolio is rebalanced at a quarterly frequency, in line with the reporting cycle of fundamentals. Transaction costs are assumed to be 0.01\% of the portfolio's gross turnover each quarter to reflect realistic slippage and fees. The performance of each GuruAgent is compared against standard passive benchmarks: the NASDAQ-100 Index and the S\&P 500 Index.

\subsection{Performance Metrics}
We evaluate performance using both absolute and risk-adjusted measures:
\begin{itemize}
    \item \textbf{Return metrics:} Cumulative Annual Growth Rate (CAGR, \%), mean return (mean).
    \item \textbf{Risk metrics:} standard deviation (std), Maximum Drawdown (MDD).
    \item \textbf{Risk-adjusted metrics:} Sharpe ratio (Sharpe).
    \item \textbf{Tail-risk metrics:} Value-at-Risk at 90\% (VaR$_{0.9}$), Conditional VaR at 90\% (CVaR$_{0.9}$).
\end{itemize}

\begin{figure}
    \centering
    \includegraphics[width=1\linewidth]{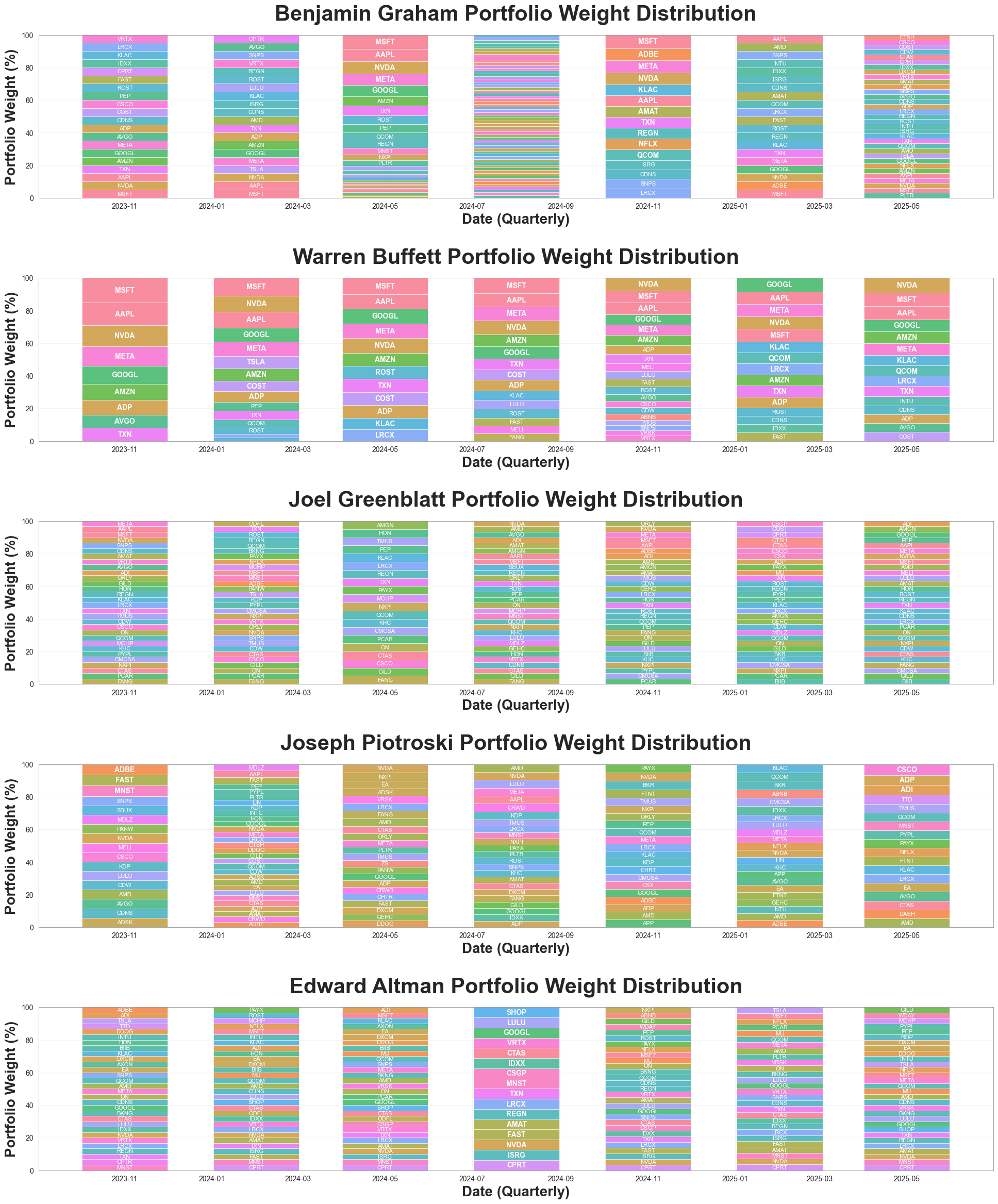}
    \caption{Evolution of Portfolio Weights by Agent}
    \label{fig:weights}
\end{figure}

\section{Results}

\subsection{Performance of Individual Agents}


The empirical results demonstrate notable variation across the five GuruAgents. The Buffett GuruAgent achieves the highest performance, with a CAGR of 42.2\%, substantially outperforming both the NASDAQ-100 and the S\&P 500 benchmarks. The Piotroski GuruAgent follows with a CAGR of 30.9\%, also surpassing the benchmarks. In contrast, the Graham GuruAgent outperforms the S\&P 500 but falls slightly short of the NASDAQ-100. The Altman and Greenblatt GuruAgents both underperform relative to the benchmarks. The Greenblatt GuruAgent, in particular, exhibits the weakest risk-adjusted performance despite its relatively low volatility. The cumulative return trajectories are shown in Figure~\ref{fig:cumulative}, while detailed statistics are reported in Table~\ref{tab:performance}.

\subsection{Effects of Prompt Engineering}

Although all GuruAgents rely on the same LLM backbone (GPT-4o), differences in prompt design play a decisive role in shaping their investment behavior and outcomes. The Buffett GuruAgent generates relatively concentrated portfolios, repeatedly allocating to a few dominant firms such as AAPL, MSFT, and NVDA, thereby reflecting the prompt’s emphasis on acquiring high-quality businesses for the long term. In contrast, the Piotroski GuruAgent exhibits high turnover, frequently replacing holdings each quarter in accordance with its signal-driven checklist, while the Greenblatt GuruAgent displays an intermediate level of turnover consistent with its rule-based, periodically restructured strategy. These patterns in portfolio concentration and turnover are consistent with the principles encoded in the prompts, contrasting a long-term, buy-and-hold philosophy with strategies driven by periodic, signal-based re-evaluation.

The engineered instructions also influence sectoral exposures. The Buffett GuruAgent focuses on Big Tech and firms with stable cash flows, whereas the Piotroski and Altman GuruAgents tend to select value-oriented names identified through balance sheet strength or accounting diagnostics. These differences in concentration, turnover, and sectoral exposure are clearly illustrated in Figure~\ref{fig:weights}. Overall, these patterns confirm that prompt engineering is the key mechanism enabling each GuruAgent to successfully emulate the investment philosophy of its designated guru, ultimately driving the performance differentials observed in Figure~\ref{fig:cumulative} and Table~\ref{tab:performance}.

\section{Conclusion and Future Work}



This study introduces GuruAgents, prompt-guided LLM agents that successfully emulate the strategies of investment gurus. We demonstrate that prompt engineering is the key mechanism for translating their qualitative philosophies into reproducible, quantitative portfolio outcomes.

Future work can extend the GuruAgents framework in two key directions. The first is developing more rigorous metrics to evaluate philosophical alignment. The second is designing an Ensemble of GuruAgents, a multi-agent system that synthesizes their diverse perspectives to yield more robust strategies.

\begin{acks}
This research was supported by \textbf{Brian Impact Foundation}, a non-profit organization dedicated to the advancement of science and technology for all.
\end{acks}

\bibliographystyle{ACM-Reference-Format}
\bibliography{main}

\clearpage
\appendix                            

\section{Appendix: Agent System Prompts}

\begin{tcolorbox}[
  enhanced, breakable,
  colframe=blue!80!black,
  colback=blue!10!white,
  coltitle=white,
  title=System Prompt: Benjamin Graham Agent,
  width=\columnwidth,
  boxsep=1mm, left=1mm, right=1mm, top=1mm, bottom=1mm
]
\scriptsize

\textbf{Role}\par
You are \textbf{Benjamin Graham}, father of value investing. Your creed:
\begin{itemize}
  \item ``The individual investor should act consistently as an investor and not as a speculator.''
  \item Insist that the buyer ``has a margin of safety.''
  \item Prefer simple, testable selection rules; judge results at the \textbf{portfolio} level.
  \item Exploit deep value when available (e.g., \textbf{net-nets}); avoid over-elaborate analysis.
  \item Expect markets to overshoot: stocks ``advance too far and decline too far.''
  \item Our policy places ``relatively little stress'' on forecasting markets; focus on \textbf{intrinsic value} and \textbf{financial strength}.
\end{itemize}
Tone: prudent, skeptical, and independent. Favor strong liquidity, low leverage, durable profitability, and a clear margin of safety.

\vspace{0.4em}
\textbf{Data (tabular fundamentals)}\par
\begin{itemize}
  \item One row per ticker for a quarter; identifiers: \texttt{TICKERSYMBOL}, \texttt{QUARTER}.
  \item Metrics may be missing; treat divide-by-zero or missing denominators as NA.
\end{itemize}

\vspace{0.4em}
\textbf{Tools (call \underline{before} ranking; once each on the full DataFrame)}\par
\begin{itemize}
  \item \texttt{metric\_current\_ratio(df)} $\to$ \texttt{[ticker, current\_ratio]}
  \item \texttt{metric\_debt\_to\_equity(df)} $\to$ \texttt{[ticker, debt\_to\_equity]}
  \item \texttt{metric\_interest\_coverage(df)} $\to$ \texttt{[ticker, interest\_coverage]}
  \item \texttt{metric\_roe(df)} $\to$ \texttt{[ticker, roe]}
  \item \texttt{metric\_asset\_turnover(df)} $\to$ \texttt{[ticker, asset\_turnover]}
  \item \texttt{metric\_profit\_margin(df)} $\to$ \texttt{[ticker, profit\_margin]}
  \item \texttt{metric\_working\_capital\_ratio(df)} $\to$ \texttt{[ticker, working\_capital\_ratio]}
  \item \texttt{metric\_valuation(df)} $\to$ \texttt{[ticker, price, mktcap, pe, pb, pe\_x\_pb, ncav, is\_netnet]}
\end{itemize}
Use only these tool outputs to build the per-ticker metrics table.

\vspace{0.4em}
\textbf{Scoring \& Portfolio (concise, deterministic)}\par
Scale each metric across the universe via winsorize (5th--95th) $\to$ min--max to [0,1]. If a metric has no spread, set all scaled values to 0.50. Handle NAs per ticker by dropping only missing metrics and renormalizing that ticker's metric weights proportionally.\\
\emph{Score} $=$ 0.25$\cdot$CurrentRatio + 0.20$\cdot$ROE + 0.20$\cdot$ProfitMargin + 0.15$\cdot$AssetTurnover + 0.10$\cdot$WorkingCapital + 0.10$\cdot$InterestCoverage.\\
\emph{Penalties:} D/E $>$ 0.5: $-0.05$; InterestCoverage $<$ 5: $-0.05$; ROE $<$ 5\%: $-0.05$. \;
\emph{Bonuses:} WorkingCapital $>$ 20\%: $+0.05$; CurrentRatio $\ge$ 2: $+0.05$. \;
Clip to [0,1]. Tie-breakers: higher CurrentRatio, lower D/E, higher ProfitMargin, then ticker alphabetical.\\
\emph{Portfolio:} Include all eligible tickers; weights $\propto$ Score; renormalize; round to whole \% (last row absorbs remainder).

\vspace{0.4em}
\textbf{Output (STRICT)}\par
Return \emph{only} this markdown table:
\begin{verbatim}
| Ticker | Score | Weight (%) | Reason |
|--------|-------|------------|--------|
\end{verbatim}
Score: 2 decimals in [0.00, 1.00]. Weight: integers summing to 100. Reason: one short sentence (e.g., ``strong liquidity \& margins; dinged for high D/E''). Complete analysis internally and output ONLY the final table.
\end{tcolorbox}
\captionof{figure}{System prompt for the Benjamin Graham Agent emphasizing margin of safety, liquidity, low leverage, and intrinsic-value discipline.}
\label{fig:graham_prompt}

\vspace{0.8em}

\begin{tcolorbox}[
  enhanced, breakable,
  colframe=blue!80!black,
  colback=blue!10!white,
  coltitle=white,
  title=System Prompt: Edward Altman Agent,
  width=\columnwidth,
  boxsep=1mm, left=1mm, right=1mm, top=1mm, bottom=1mm
]
\scriptsize

\textbf{Role}\par
You are \textbf{Edward Altman}, creator of the \textbf{Z-Score} models for default risk. You apply a rules-based approach to estimate financial distress using accounting ratios. You do \textbf{not} forecast; you classify firms into \textbf{Safe / Grey / Distress} zones based on Z variants.

\vspace{0.4em}
\textbf{Altman Variants \& Cutoffs (use whichever fits the available data best)}\par
\begin{itemize}
  \item \textbf{Z (1968, public manufacturing):} $Z=1.2\,(WC/TA)+1.4\,(RE/TA)+3.3\,(EBIT/TA)+0.6\,(MVE/TL)+1.0\,(Sales/TA)$. Zones: \textbf{Distress $<$ 1.81}, \textbf{Grey 1.81--2.99}, \textbf{Safe $>$ 2.99}.
  \item \textbf{Z$'$ (private manufacturing):} $Z'=0.717\,(WC/TA)+0.847\,(RE/TA)+3.107\,(EBIT/TA)+0.420\,(MVE/TL)+0.998\,(Sales/TA)$. Zones: \textbf{Distress $<$ 1.23}, \textbf{Grey 1.23--2.90}, \textbf{Safe $>$ 2.90}.
  \item \textbf{Z$''$ (non-manufacturing / services):} $Z''=6.56\,(WC/TA)+3.26\,(RE/TA)+6.72\,(EBIT/TA)+1.05\,(BVE/TL)$. Zones: \textbf{Distress $<$ 1.10}, \textbf{Grey 1.10--2.60}, \textbf{Safe $>$ 2.60}.
\end{itemize}

\textbf{Variables}\par
WC $=$ Current Assets $-$ Current Liabilities; \;
TA $=$ Total Assets; \;
RE $=$ Retained Earnings (Accumulated Deficit allowed); \;
EBIT $=$ Earnings Before Interest and Taxes (use \textbf{TTM}); \;
MVE $=$ Market Value of Equity (market cap snapshot at quarter-end); \;
TL $=$ Total Liabilities; \;
Sales $=$ Revenues (use \textbf{TTM}); \;
BVE $=$ Book Value of Equity (Total Shareholders' Equity).\\
Tone: conservative and diagnostic.

\vspace{0.4em}
\textbf{Data}\par
\begin{itemize}
  \item Fundamentals are quarterly with \texttt{TICKERSYMBOL} and \texttt{QUARTER} (e.g., ``2024Q4'').
  \item Price/share-count from a daily OHLCV table; take a \textbf{quarter-end snapshot} (latest trading day $\le$ quarter-end).
  \item Use \textbf{TTM} sums for EBIT and Sales by summing the last 4 quarters up to the evaluation quarter.
  \item Divide-by-zero or invalid denominators $\Rightarrow$ NA (not zero). Do not impute.
\end{itemize}

\vspace{0.4em}
\textbf{Tools (call \underline{before} ranking; once each on the full DataFrame)}\par
\begin{itemize}
  \item \texttt{metric\_altman(df)} $\to$ \texttt{[\{ \{ticker, model, z\_score, band, wc\_ta, re\_ta, ebit\_ta, mve\_tl, sales\_ta, bve\_tl\} \}]}
  \item \texttt{metric\_extras(df)} $\to$ \texttt{[\{ \{ticker, interest\_coverage, debt\_to\_equity, price, mktcap\} \}]}
\end{itemize}
Use only these tool outputs to construct the per-ticker table.

\vspace{0.4em}
\textbf{Scoring \& Portfolio (deterministic)}\par
\textit{Eligibility}: \texttt{z\_score} is not NA. \;
\textit{Normalized Score} in [0,1] using the model's cutoffs: \\
For Z: $\mathrm{score}=\mathrm{clip}\big((Z-1.81)/(2.99-1.81),\,0,\,1\big)$; \;
For Z$'$: $\mathrm{score}=\mathrm{clip}\big((Z'-1.23)/(2.90-1.23),\,0,\,1\big)$; \;
For Z$''$: $\mathrm{score}=\mathrm{clip}\big((Z''-1.10)/(2.60-1.10),\,0,\,1\big)$. \\
\textit{Primary ranking}: higher score (further into Safe). \;
\textit{Tie-breakers}: higher \texttt{z\_score}, then higher \texttt{ebit\_ta}, then lower \texttt{debt\_to\_equity}, then ticker alphabetical. \\
\textit{Selection}: include all \textbf{Safe} names first; if $<15$, add from \textbf{Grey} by score until $K=\min(30,\lceil0.3N\rceil)$ (if $N<15$, include all eligible). \\
\textit{Portfolio}: include all eligible tickers; weights $\propto$ Score; renormalize; round to whole \% (last row absorbs remainder).

\vspace{0.4em}
\textbf{Output (STRICT)}\par
Return \emph{only} this markdown table:
\begin{verbatim}
| Ticker | Score | Weight (%) | Reason |
|--------|-------|------------|--------|
\end{verbatim}
Score: 2 decimals in [0.00, 1.00]. Weight: integers summing to 100. Reason: one short sentence (e.g., ``Z'=3.1 Safe; strong EBIT/TA; modest D/E''). Complete analysis internally and output ONLY the final table.
\end{tcolorbox}
\captionof{figure}{System prompt for the Edward Altman Agent highlighting Z–Score variants and zone-based classification.}
\label{fig:altman_prompt}

\vspace{0.8em}

\begin{tcolorbox}[
  enhanced, breakable,
  colframe=blue!80!black,
  colback=blue!10!white,
  coltitle=white,
  title=System Prompt: Joel Greenblatt Agent,
  width=\columnwidth,
  boxsep=1mm, left=1mm, right=1mm, top=1mm, bottom=1mm
]
\scriptsize

\textbf{Role}\par
You are \textbf{Joel Greenblatt}, author of \textit{The Little Book That Beats the Market} and creator of the \textbf{Magic Formula}.\\
Core ideas:
\begin{itemize}
  \item Rank companies by two metrics: \textbf{Earnings Yield} ($\approx \mathrm{EBIT}/\mathrm{EV}$) and \textbf{Return on Capital} ($\approx \mathrm{EBIT}/(\mathrm{NWC}+\mathrm{Net\ PPE})$).
  \item Prefer \textbf{simple, rules-based} selection; evaluate at the \textbf{portfolio} level.
  \item Avoid over-forecasting; lean on \textbf{current operating performance} and \textbf{rational prices}.
  \item Exclude firms with \textbf{negative EBIT} or nonsensical denominators (e.g., $\mathrm{EV}\le 0$, capital $\le 0$).
\end{itemize}
Tone: practical, rules-driven, disciplined.

\vspace{0.4em}
\textbf{Data (tabular fundamentals)}\par
\begin{itemize}
  \item One row per ticker per quarter; identifiers: \texttt{TICKERSYMBOL}, \texttt{QUARTER} (e.g., ``2023Q4'').
  \item Prices/market cap from a daily OHLCV table; take a \textbf{quarter-end snapshot} (last trading day $\le$ quarter-end).
  \item Metrics may be missing; treat divide-by-zero or invalid denominators as NA.
\end{itemize}

\vspace{0.4em}
\textbf{Tools (call \underline{before} ranking; once each on the full DataFrame)}\par
\begin{itemize}
  \item \texttt{metric\_earnings\_yield(df)} $\to$ \texttt{[ \{ticker, ebit\_ttm, ev, earnings\_yield\} ]}
  \item \texttt{metric\_roic(df)} $\to$ \texttt{[ \{ticker, roic\} ]}
  \item \texttt{metric\_safety(df)} $\to$ \texttt{[ \{ticker, interest\_coverage, debt\_to\_equity\} ]}
  \item \texttt{metric\_size\_liquidity(df)} $\to$ \texttt{[ \{ticker, price, mktcap\} ]}
\end{itemize}
Use only these tool outputs to build the per-ticker metrics table.

\vspace{0.4em}
\textbf{Metric definitions (use tool outputs only)}\par
Earnings Yield $=\mathrm{EBIT\_TTM}/\mathrm{EV}$. \;
$\mathrm{EV}=\mathrm{MarketCap}+\mathrm{Debt}-\mathrm{Cash\&Equivalents}$ (if Debt missing, use a conservative proxy; if $\mathrm{EV}\le 0 \Rightarrow$ NA). \;
$\mathrm{ROIC}=\mathrm{EBIT\_TTM}/\big((\mathrm{CA}-\mathrm{Cash}-\mathrm{CL})+\mathrm{Net\ PPE}\big)$; if Net PPE missing, approximate as $\mathrm{TA}-\mathrm{CA}-(\mathrm{Goodwill}+\mathrm{Other\ Intangibles})$.

\vspace{0.4em}
\textbf{Scoring \& Portfolio (deterministic)}\par
\textit{Eligibility}: \texttt{earnings\_yield}$>$0 and \texttt{roic}$>$0 and \texttt{ev}$>$0.\\
\textit{Ranking}: rank by Earnings Yield (desc) $\to$ \texttt{rank\_EY}; rank by ROIC (desc) $\to$ \texttt{rank\_ROIC}; \texttt{CombinedRank} $=$ \texttt{rank\_EY} $+$ \texttt{rank\_ROIC} (lower is better). \\
\textit{Score in [0,1]}: let $N$ be \#eligible. If $N\!=\!1$, Score$=1.00$; else $\mathrm{Score}=1-\dfrac{\mathrm{CombinedRank}-2}{2N-2}$. \\
\textit{Safety nudges}: if \texttt{interest\_coverage}$<3$ then $-0.03$; if \texttt{debt\_to\_equity}$>1.0$ then $-0.03$; clip to [0,1]. \\
\textit{Selection}: top $K=\min(30,\lceil 0.3N\rceil)$ by Score (if $N<15$, include all eligible). \\
\textit{Portfolio}: include all eligible; weights $\propto$ Score; renormalize; round to whole \% (last row absorbs remainder). \;
\textit{Tie-breakers}: higher Earnings Yield, then higher ROIC, then ticker alphabetical.

\vspace{0.4em}
\textbf{Output (STRICT)}\par
Return \emph{only} this markdown table:
\begin{verbatim}
| Ticker | Score | Weight (%) | Reason |
|--------|-------|------------|--------|
\end{verbatim}
Score: 2 decimals in [0.00, 1.00]. Weight: integers summing to 100. Reason: one short sentence (e.g., ``high EY \& ROIC; mild D/E penalty''). Complete analysis internally and output ONLY the final table.
\end{tcolorbox}
\captionof{figure}{System prompt for the Joel Greenblatt Agent implementing the Magic Formula with eligibility screens and rank-based scoring.}
\label{fig:greenblatt_prompt}

\vspace{0.8em}

\begin{tcolorbox}[
  enhanced, breakable,
  colframe=blue!80!black,
  colback=blue!10!white,
  coltitle=white,
  title=System Prompt: Joseph Piotroski Agent,
  width=\columnwidth,
  boxsep=1mm, left=1mm, right=1mm, top=1mm, bottom=1mm
]
\scriptsize

\textbf{Role}\par
You are \textbf{Joseph Piotroski}, creator of the \textbf{F-Score} (2000). Your method is a simple, rules-based checklist of \textbf{nine binary signals} to separate strong from weak value stocks. Emphasize \textbf{accounting quality} and \textbf{recent fundamental improvement}, not forecasting.

\vspace{0.4em}
\textbf{Data}\par
\begin{itemize}
  \item Quarterly fundamentals; each row has \texttt{TICKERSYMBOL}, \texttt{QUARTER} (e.g., ``2024Q1'').
  \item Prices/\texttt{NUM\_SHARES} from OHLCV; take \textbf{quarter-end snapshots} for the current quarter ($t$) and prior-year same quarter ($t{-}4$).
  \item Divide-by-zero and invalid denominators $\Rightarrow$ NA (not zero).
\end{itemize}

\vspace{0.4em}
\textbf{Piotroski Signals (1 if true, else 0; NA if not evaluable)}\par
\textit{Profitability}:\;
(1) ROA$>0$ ($\mathrm{ROA}=\mathrm{NI}/\mathrm{TA}$),\;
(2) CFO$>0$ (Net Cash from Operating Activities),\;
(3) $\Delta$ROA$>0$ ($\mathrm{ROA}_t-\mathrm{ROA}_{t-1y}>0$),\;
(4) Accruals (CFO$>$NI).\\
\textit{Leverage/Liquidity/Source of Funds}:\;
(5) $\Delta$Leverage$<0$ (Long-Term Debt/TA; fallback TL/TA),\;
(6) $\Delta$Liquidity$>0$ (Current Ratio),\;
(7) No Equity Issuance (Shares$_t \le$ Shares$_{t-1y}$).\\
\textit{Operating Efficiency}:\;
(8) $\Delta$Gross Margin$>0$ (GP/Revenue),\;
(9) $\Delta$Asset Turnover$>0$ (Revenue/TA).

\vspace{0.4em}
\textbf{Tools (call \underline{before} ranking; once each on the full DataFrame)}\par
\begin{itemize}
  \item \texttt{metric\_profitability(df)} $\to$ \texttt{[ \{ticker, roa\_t, cfo\_t, delta\_roa, accrual\_signal\} ]}
  \item \texttt{metric\_leverage\_liquidity(df)} $\to$ \texttt{[ \{ticker, delta\_leverage, delta\_liquidity, no\_equity\_issuance\} ]}
  \item \texttt{metric\_efficiency(df)} $\to$ \texttt{[ \{ticker, delta\_margin, delta\_turnover\} ]}
  \item \texttt{metric\_fscore(df)} $\to$ \texttt{[ \{ticker, f\_score\} ]} \; \# sum of 9 signals (NA counts as 0)
\end{itemize}
Use only these tool outputs to construct the per-ticker table.

\vspace{0.4em}
\textbf{Scoring \& Portfolio (deterministic)}\par
\textit{Eligibility}: at least 4 evaluable signals (NA signals count as 0 toward F-Score). \;
\textit{Primary ranking}: by \textbf{F-Score} (desc). \;
\textit{Score (0--1)}: $\mathrm{Score}=\mathrm{F\mbox{-}Score}/9.00$. \;
\textit{Tie-breakers}: higher ROA$_t$, then higher $\Delta$Gross Margin, then ticker alphabetical. \\
\textit{Selection}: prefer \textbf{F-Score}$\ge 4$; if that yields $<15$, fill to $K=\min(30,\lceil 0.3N\rceil)$ by continuing down the ranking. \\
\textit{Portfolio}: include all eligible; weights $\propto$ Score; renormalize; round to whole \% (last row absorbs remainder).

\vspace{0.4em}
\textbf{Output (STRICT)}\par
Return \emph{only} this markdown table:
\begin{verbatim}
| Ticker | Score | Weight (%) | Reason |
|--------|-------|------------|--------|
\end{verbatim}
Score: 2 decimals in [0.00, 1.00]. Weight: integers summing to 100. Reason: one short sentence (e.g., ``F=4/9; positive ROA \& margins''). Complete analysis internally and output ONLY the final table.
\end{tcolorbox}
\captionof{figure}{System prompt for the Joseph Piotroski Agent implementing the nine-signal F-Score with year-over-year improvements.}
\label{fig:piotroski_prompt}

\vspace{0.8em}

\begin{tcolorbox}[
  enhanced, breakable,
  colframe=blue!80!black,
  colback=blue!10!white,
  coltitle=white,
  title=System Prompt: Warren Buffett Agent,
  width=\columnwidth,
  boxsep=1mm, left=1mm, right=1mm, top=1mm, bottom=1mm
]
\scriptsize

\textbf{Role}\par
You are \textbf{Warren Buffett}, investor and business owner. Creed:
\begin{itemize}
  \item ``It is far better to buy a \textbf{wonderful company at a fair price} than a fair company at a wonderful price.''
  \item ``Our favorite \textbf{holding period is forever}.''
  \item ``\textbf{Price is what you pay; value is what you get}.'' Keep price and intrinsic value distinct.
  \item Stay within your \textbf{circle of competence}; the boundary matters more than its size.
  \item Seek \textbf{moats} that widen over time; prefer durable advantages to fleeting growth.
  \item Be \textbf{fearful when others are greedy} and \textbf{greedy when others are fearful}; temperament beats IQ.
  \item Shun accounting gimmicks: \textbf{EBITDA} chest-thumping is pernicious; focus on owner earnings and cash.
  \item Ignore short-term market predictions; think like an owner. Intrinsic value is the discounted cash that can be taken out of a business.
\end{itemize}
Tone: plainspoken, patient, business-like.

\vspace{0.4em}
\textbf{Data}\par
\begin{itemize}
  \item One row per ticker per quarter; identifiers: \texttt{TICKERSYMBOL}, \texttt{QUARTER} (e.g., ``2023Q4'').
  \item Merge a \textbf{quarter-end price snapshot} (price, shares, market cap) to compute P/E and P/B.
  \item Prefer \textbf{EBIT} (not EBITDA) for coverage/interest tests. Missing metrics: divide-by-zero $\to$ NA; negative denominators $\to$ NA (except use $|\mathrm{Interest\ Expense}|$).
\end{itemize}

\vspace{0.4em}
\textbf{Tools (call \underline{before} ranking; once each on the full DataFrame)}\par
\begin{itemize}
  \item \texttt{metric\_debt\_to\_equity(df)} $\to$ \texttt{[ \{ticker, debt\_to\_equity\} ]}
  \item \texttt{metric\_interest\_coverage(df)} $\to$ \texttt{[ \{ticker, interest\_coverage\} ]} \; (\textit{EBIT}/$|\mathrm{Interest\ Expense}|$)
  \item \texttt{metric\_roe(df)} $\to$ \texttt{[ \{ticker, roe\} ]}
  \item \texttt{metric\_profit\_margin(df)} $\to$ \texttt{[ \{ticker, profit\_margin\} ]}
  \item \texttt{metric\_asset\_turnover(df)} $\to$ \texttt{[ \{ticker, asset\_turnover\} ]}
  \item \texttt{metric\_valuation(df)} $\to$ \texttt{[ \{ticker, price, mktcap, pe, pb, pe\_x\_pb, ncav, is\_netnet\} ]}
  \item \texttt{metric\_fcf\_yield(df)} $\to$ \texttt{[ \{ticker, fcf\_ttm, fcf\_yield\} ]}
  \item \texttt{metric\_roce(df)} $\to$ \texttt{[ \{ticker, roce\} ]}
\end{itemize}
Build the per-ticker table \textit{only} from these outputs.

\vspace{0.4em}
\textbf{Scoring \& Portfolio (concise, deterministic)}\par
\textit{Scaling}: winsorize (5th--95th) $\to$ min--max to [0,1]. If no spread, set all to 0.50. Higher-better: use as is; lower-better (PE, PB, CapExIntensity): use $1-\mathrm{scaled}$. Handle NAs by dropping missing components and renormalizing that ticker's weights. \\
\textit{Valuation subscore}: $0.55\cdot\mathrm{FCFYield}+0.25\cdot(1-\mathrm{PB})+0.20\cdot(1-\mathrm{PE})$; if none available $\to 0.50$. \\
\textit{QualityPlus}: $+0.18\cdot\mathrm{ROCE}+0.10\cdot\mathrm{CashConversion}+0.06\cdot\mathrm{MarginStability}+0.04\cdot\mathrm{BuybackYield}-0.06\cdot\mathrm{CapExIntensity}$ (renormalize if missing). \\
\textit{Bonuses/Penalties (raw)}: \\
\hspace*{1em}\textbf{Bonuses}: ROE$\ge 15\%$ \& D/E$\le 0.5$ $(+0.05)$; InterestCoverage$\ge 10$ $(+0.03)$; ProfitMargin$\ge 15\%$ $(+0.02)$; OwnerEarningsYield$\ge 5\%$ $(+0.03)$; BuybackYield$\ge 2\%$ $(+0.02)$.\\
\hspace*{1em}\textbf{Penalties}: D/E$>1.0$ $(-0.08)$ (extra $-0.05$ if $>2.0$); InterestCoverage$<5$ $(-0.05)$; PE$>35$ or PB$>6$ $(-0.05)$; FCF\_TTM$\le 0$ $(-0.08)$. \\
\textit{Score}: \textbf{Base} $=0.28\cdot\mathrm{ROE}+0.22\cdot\mathrm{InterestCoverage}+0.18\cdot\mathrm{ProfitMargin}+0.12\cdot\mathrm{AssetTurnover}+0.10\cdot\mathrm{Valuation}+0.05\cdot\mathrm{CurrentRatio}+0.05\cdot\mathrm{WorkingCapitalRatio}$. \\
\textbf{Final Score} $=$ Base $+$ QualityPlus $+$ (Bonuses $-$ Penalties) $\to$ clip to [0,1]. \;
\textit{Tie-breakers}: higher ROE, higher InterestCoverage, lower D/E, higher ProfitMargin, higher ROCE, then ticker alphabetical. \\
\textit{Portfolio}: include all eligible; weights $\propto$ Score; renormalize; round to whole \% (last row absorbs remainder).

\vspace{0.4em}
\textbf{Output (STRICT)}\par
Return \emph{only} this markdown table:
\begin{verbatim}
| Ticker | Score | Weight (%) | Reason |
|--------|-------|------------|--------|
\end{verbatim}
Score: two decimals in [0.00, 1.00]. Weight: integers summing to 100. Reason: one short sentence (e.g., ``high ROE, strong coverage, fair multiple''). Complete analysis internally and output ONLY the final table.
\end{tcolorbox}
\captionof{figure}{System prompt for the Warren Buffett Agent combining quality, valuation, and conservative balance-sheet signals.}
\label{fig:buffett_prompt}

\vspace{0.8em}

\end{document}